\documentclass[sigconf]{acmart}
\AtBeginDocument{%
  }

\setcopyright{acmlicensed}
\copyrightyear{2026}
\acmYear{2026}
\setcopyright{cc}
\setcctype{by}
\acmConference[WWW '26]{Proceedings of the ACM Web Conference 2026}{April 13--17, 2026}{Dubai, United Arab Emirates}
\acmBooktitle{Proceedings of the ACM Web Conference 2026 (WWW '26), April 13--17, 2026, Dubai, United Arab Emirates}
\acmPrice{}
\acmDOI{10.1145/3774904.3792545}
\acmISBN{979-8-4007-2307-0/2026/04}

\usepackage{multirow}
\usepackage{textcomp}
\usepackage{pifont}
\usepackage{booktabs}
\usepackage{amsmath}
\usepackage{amsfonts}
\usepackage{subfigure}
\usepackage{algorithm}
\usepackage{algorithmic}
\usepackage {balance}

\begin{document}

\title{ContiGuard: A Framework for Continual Toxicity Detection Against Evolving Evasive Perturbations}


\settopmatter{authorsperrow=4}

\author{Hankun Kang}
\orcid{https://orcid.org/0009-0006-0780-1339}
\affiliation{%
  \department{School of Computer Science}
  \institution{Wuhan University}
  \city{Wuhan}
  \state{Hubei}
  \country{China}
}
\email{kanghankun@whu.edu.cn}

\author{Xin Miao}
\orcid{https://orcid.org/0000-0001-6689-0606}
\affiliation{%
	\department{School of Computer Science}
	\institution{Wuhan University}
	\city{Wuhan}
	\state{Hubei}
	\country{China}
}
\email{miaoxin@whu.edu.cn}

\author{Jianhao Chen}
\orcid{https://orcid.org/0009-0005-1907-8988}
\affiliation{%
	\department{School of Computer Science}
	\institution{Wuhan University}
	\city{Wuhan}
	\state{Hubei}
	\country{China}
}
\affiliation{%
  \institution{Zhongguancun Academy}
  \city{Beijing}
  \country{China}
}
\email{chenjianhao@whu.edu.cn}

\author{Jintao Wen}
\orcid{https://orcid.org/0000-0001-6355-3014}
\affiliation{%
	\department{School of Computer Science}
	\institution{Wuhan University}
	\city{Wuhan}
	\state{Hubei}
	\country{China}
}
\email{23jtwen@whu.edu.cn}

\author{Mayi Xu}
\orcid{https://orcid.org/0000-0002-1877-224X}
\affiliation{%
	\department{School of Computer Science}
	\institution{Wuhan University}
	\city{Wuhan}
	\state{Hubei}
	\country{China}
}
\email{xumayi@whu.edu.cn}

\author{Weiyu Zhang}
\orcid{https://orcid.org/0000-0002-4646-1991}
\authornotemark[1]
\affiliation{%
  \department{Key Laboratory of Computing Power Network and Information Security, Ministry of Education, Shandong Computer Science Center (National Supercomputer Center in Jinan)}
  \institution{Qilu University of Technology (Shandong Academy of Sciences)}
  \city{Jinan}
  \state{Shandong}
  \country{China}
}
\affiliation{%
	\department{Shandong Provincial Key Laboratory of Computing Power Internet and Service Computing}
	\institution{Shandong Fundamental Research Center for Computer Science}
	\city{Jinan}
	\state{Shandong}
	\country{China}
}
\email{zwy@qlu.edu.cn}

\author{Wenpeng Lu}
\orcid{https://orcid.org/0000-0002-1840-3540}
\authornotemark[1]
\affiliation{%
	\department{Key Laboratory of Computing Power Network and Information Security, Ministry of Education, Shandong Computer Science Center (National Supercomputer Center in Jinan)}
	\institution{Qilu University of Technology (Shandong Academy of Sciences)}
	\city{Jinan}
	\state{Shandong}
	\country{China}
}
\affiliation{%
	\department{Shandong Provincial Key Laboratory of Computing Power Internet and Service Computing}
	\institution{Shandong Fundamental Research Center for Computer Science}
	\city{Jinan}
	\state{Shandong}
	\country{China}
}
\email{wenpeng.lu@qlu.edu.cn}

\author{Tieyun Qian}
\orcid{https://orcid.org/0000-0003-4667-5794}
\authornote{Corresponding authors.}
\affiliation{%
	\department{School of Computer Science}
	\institution{Wuhan University}
	\city{Wuhan}
	\state{Hubei}
	\country{China}
}
\affiliation{%
  \institution{Zhongguancun Academy}
  \city{Beijing}
  \country{China}
}
\email{qty@whu.edu.cn}

\renewcommand{\shortauthors}{Hankun Kang et al.}

\begin{abstract}
    Toxicity detection mitigates the dissemination of toxic content (e.g., hateful comments, posts, and messages within online social actions) to safeguard a healthy online social environment. However, malicious users persistently develop evasive perturbations to disguise toxic content and evade detectors. Traditional detectors or methods are static over time and are inadequate in addressing these evolving evasion tactics. Thus, continual learning emerges as a logical approach to dynamically update detection ability against evolving perturbations.
    Nevertheless, disparities across perturbations hinder the detector’s continual learning on perturbed text. More importantly, perturbation-induced noises distort semantics to degrade comprehension and also impair critical feature learning to render detection sensitive to perturbations. These amplify the challenge of continual learning against evolving perturbations.
    
    In this work, we present ContiGuard, the first framework tailored for continual learning of the detector on time-evolving perturbed text (termed continual toxicity detection) to enable the detector to continually update capability and maintain sustained resilience against evolving perturbations. 
    Specifically, to boost the comprehension, we present an LLM powered semantic enriching strategy, where we dynamically incorporate possible meaning and toxicity-related clues excavated by LLM into the perturbed text to improve the comprehension. To mitigate non-critical features and amplify critical ones, we propose a discriminability driven feature learning strategy, where we strengthen discriminative features while suppressing the less-discriminative ones to shape a robust classification boundary for detection. Additionally, we introduce a historical capability replay strategy to preserve previously learned features via feature alignment to alleviate capability forgetting. To the best of our knowledge, this work is the first study on continual toxicity detection against time-evolving evasive perturbed text. Extensive experiments prove the superior performance of ContiGuard over both existing detectors and continual methods.
    Code and dataset are available at \url{https://github.com/khk-abc/ContiGuard}.
    
    \textit{\textbf{Warning}: This paper contains discussions of harmful content that may be disturbing to some readers.}
    
\end{abstract}

%




\maketitle

\section{Introduction}

Toxicity detection aims to identify toxic text like hateful content~\cite{fortuna2018survey}, social bias~\cite{sap2020social}, and stereotypes~\cite{gongane2022detection}, to safeguard healthy social interactions, which is critical for society safety~\cite{dixon2018measuring}. However, detection efforts face significant challenges due to the continuous evolution of evasive tactics: malicious users regularly create text perturbations to bypass detectors, such as character repetition (\textit{iiiddioot}), homoglyph substitution (\textit{id10t}), and other potential tactics.

Existing detection methods struggle to adapt to the continuous evolution of evasive perturbations. Most of them are tailored for ordinary text (e.g., idiot), limiting their ability to identify perturbed toxic text~\cite{costa2024mutox,kebriaei2024persian}.
A few methods incorporate specific types of perturbed text into training~\cite{costa2024mutox,kebriaei2024persian}, excelling at detecting known perturbations such as character repetition. But they struggle to deal with continually appearing perturbations since they do not update detection ability against appearing perturbations over time, like homoglyph substitution.
Consequently, the challenge of continually detecting emerging perturbed toxic text remains an open but overlooked problem, posing risks to society's safety.

Considering continual learning can update the detector's capability over time, we model different types of perturbed text as the ones distributed in distinct perturbation domains, and we harness domain continual learning to update the detection ability of the detector, enabling the detector to evolve along with constantly appearing perturbations and making it suitable for the evolving evasive perturbed text.

While continual learning is a logical approach to address evolving perturbed text in toxicity detection, it remains hindered by several critical challenges. 
Specifically, the heterogeneity of perturbations (e.g., \textit{iiiddioot} and \textit{id10t}) leads to the forgetting of the detector’s historical detection capability over time. Furthermore, perturbations disrupt the original textual structure, making the detector struggle to comprehend text, especially the hidden toxicity within text. In addition, the detector learns noisy or irrelevant features derived from perturbations, such as manipulation variations of perturbation (e.g., \textit{iiiddioot} vs. \textit{idiiiiottt}, both perturbed from \textit{idiot}). These features are noisy and redundant, less critical for toxicity classification, and even obscure the learning of critical toxicity-related features, rendering the detector dependent on these features and sensitive to diverse perturbed text. In summary, the perturbations corrupt the original semantics, causing the detector to struggle to understand text. They also introduce less-critical features, which occupy the detector's representational capacity and deviate the detector's focus from critical features, degrading continual detection against diverse perturbed text.

In this work, we propose ContiGuard, the first framework to continually identify evolving perturbed toxic text.
Firstly, to improve the text comprehension hindered by corrupted semantics, we propose an \textit{LLM powered semantic enriching strategy}, where we analyze that LLM can be utilized to reason auxiliary information from its parametric knowledge to enrich the insights into perturbed text, enhancing the text comprehension for detection. During the incorporation of the information into perturbed text, it also acts as a momentum factor to mitigate the risk of trapping into local optima. Specifically, we dynamically incorporate possible meaning and toxicity-related clues excavated by LLM into perturbed text to boost the comprehension.

To mitigate the interference of non-critical features and focus more on critical ones to make the detector robust to diverse perturbed text, we propose a \textit{discriminability driven feature learning strategy} since the discriminability is inherently critical to shape a reliable boundary for the robust classification, where we measure each feature's contribution using attribution analysis~\cite{sundararajan2017axiomatic} and differentiate their discriminative power. We then strengthen the discriminative features via global rotation and suppress less discriminative ones by unlearning, forcing the detector to learn core features for the robust detection.

To alleviate the forgetting of detection capability due to the lost historical features caused by the gaps among perturbations, we further adopt a \textit{historical capability replay strategy}, where we align features of memory samples between the old and current detectors to preserve historical features, maintaining the detector’s prior detection capability.

Our contributions are summarized as follows.
\begin{itemize}
    \item We first highlight a crucial yet overlooked continual toxicity detection problem, i.e., enabling the detector to continually update its detection capability along with evolving, evasive perturbed text prevalent in real-world scenarios, which is essential for safeguarding social safety.
    \item We propose ContiGuard, the first framework for continual toxicity detection, performing LLM powered semantic enriching, discriminability driven feature learning, and historical capability replay to address the key challenges.
    \item Extensive experiments show the superior performance of ContiGuard against evolving perturbed text compared to existing detectors, static methods, and continual approaches.
\end{itemize}

\begin{figure*}[!htb]
    \centering
    \includegraphics[width=0.95\textwidth]{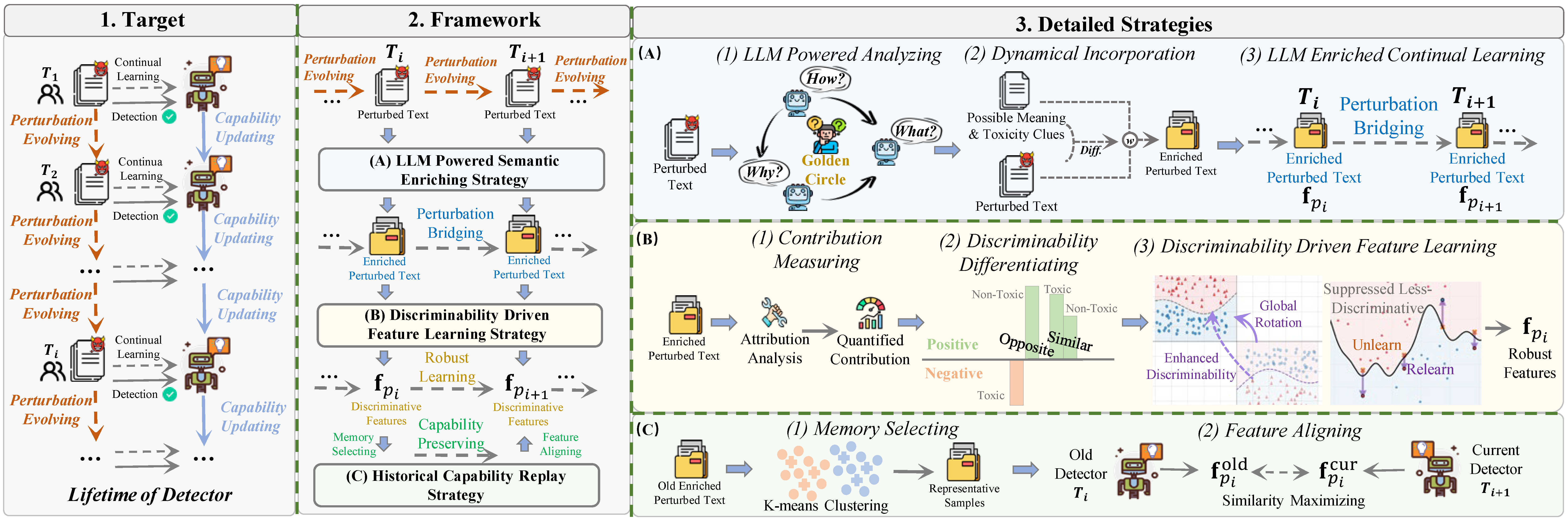}
    \caption{Illustration of our ContiGuard framework.}
    \label{fig:framework}
\end{figure*}

\section{Methodology}

\subsection{Problem Formulation}
Perturbed text is modeled as the text distributed in perturbation domains. Correspondingly, continual toxicity detection against time-evolving perturbed text is formulated as domain-incremental toxicity classification with incremental perturbation domains.
Specifically, perturbation domain $p_i$ appears at the i-th moment within the lifespan $T$ of a lifelong detector $F$. Given a perturbed toxicity dataset $D_{p_i}$ comprising $N_{p_i}$ samples $(x_{p_i,j}, y_{p_i,j})$ where $j = \{1, 2, \ldots, N_{p_i}\}$, $x_{p_i,j}$ denotes the text perturbed by $p_i$, and $y_{p_i,j} \in \mathcal{Y}$ with $\mathcal{Y} = \{0, 1\}$ is the toxicity label.
The task optimizes the parameters $\theta_F$ by minimizing the loss to maximize the conditional probabilities of labels, enabling the detector to continually recognize toxicity in evolving perturbed text over time as follows:
\begin{equation}
    \theta^*_F := \arg \max_{\theta_F} \prod_{i}^{T}\prod_{j}^{N_{p_i}} 
    p_{\theta_F}(y_{p_i,j} |x_{p_i,j}),
\label{eq:loss mini}
\end{equation}
where $p_{\theta_F}(y_{p_i,j} |x_{p_i,j})$ denotes the conditional probability of the true label computed by the detector.

\subsection{ContiGuard Framework}
As Fig.~\ref{fig:framework} shows, ContiGuard mainly includes three strategies: 

(1) \textit{LLM powered semantic enriching strategy.} Evasive perturbations intentionally obfuscate text, rendering toxicity implicit and impeding text comprehension. To address this, we analyze that LLM can provide auxiliary information to enrich insights into perturbed text for toxicity detection and can smooth gradients during optimization. Correspondingly, we leverage LLM to capture potential original meanings and toxicity-related clues and we integrate them into perturbed text to enhance comprehension.

(2) \textit{discriminability driven feature learning strategy.} Perturbations induce noises to make the features of perturbed text redundant and noisy, causing non-critical features to occupy the detector’s representation capacity and rendering detection sensitive to various perturbations. Considering the discriminability is the critical property to shape a reliable classification boundary for robust toxicity detection, we measure the discriminative power of features for toxicity detection, and we strengthen the discriminative features by global flipping and suppress less discriminative ones, guiding the detector to prioritize learning critical features.

(3) \textit{Historical capability replay strategy.} Various types of perturbations are significantly distinct so the historical detection capability against previous types of perturbed text may be lost. Hence, we retain the detector's historical capability by aligning the features of memory samples encoded with the old and current detectors.

\subsubsection{LLM Powered Semantic Enriching Strategy}
Given the rich knowledge and reasoning ability of LLM, this strategy use LLM to mine useful auxiliary information to improve the understanding of perturbed text, such as possible meaning and toxicity-related clues.

(1) \textit{Preliminaries} We first analyze how the auxiliary information from LLM enriches the perturbed text.
Let $X_{p_i}$, $X_{a_i}$, and $Y_{p_i}$ denote the perturbed text, LLM generated auxiliary information, and predicted label, respectively. The conditional probability of $Y_{p_i}$ given $X_{p_i}$ is decomposed as:
\begin{equation}
    p(Y_{p_i}|X_{p_i}) = \int p(Y_{p_i}|X_{a_i}=x_{a_i}, X_{p_i}) p(X_{a_i}=x_{a_i}|X_{p_i}) \mathrm{d}x_{a_i},
\end{equation}
where $x_{a_i}$ is the specific values of $X_{a_i}$. Hence, the objective of continual toxicity detection, maximizing $\prod_i^Tp_{\theta_F}(Y_{p_i}|X_{p_i})$, becomes:
\begin{equation}
    \max_{\theta_F} \prod_i^T\int p_{\theta_F}(Y_{p_i}|x_{a_i}, X_{p_i}) p_{\theta_{LLM}}(x_{a_i}|X_{p_i}) \mathrm{d}x_{a_i}
\end{equation}
Based on $p_{\theta_F}(Y_{p_i}|x_{a_i}, X_{p_i})$, the auxiliary information enriches the insights into perturbed text and bridges the learning on different perturbed text over time via the shared insights (e.g., the same restored terms and similar clues within various perturbations). 
In addition, the LLM parameters $\theta_{LLM}$ remain fixed during optimization to render $p_{\theta_{LLM}}(x_{a_i}|X_{p_i})$ an optimization-independent but input-dependent factor, which encapsulates the probability tendency of the toxicity-related knowledge in the LLM to adjust the learning by weighting the detector's output probability.

Furthermore, perturbations introduce noises, resulting in a jagged optimization process that is highly prone to converging to local optima. In contrast, the auxiliary information structured via ordinary text is less noisy and more flat, and the gradient derived from such information can act as a momentum term to mitigate the risk of the optimization process becoming trapped in local optima.
Specifically, we incorporate the features of auxiliary information $X_{a_i}$ into the features of perturbed text $X_{p_i}$ by linearly weighting as:
\begin{equation}
    f([X_{p_i},X_{a_i}])=\alpha f(X_{p_i})+(1-\alpha) f(X_{a_i}),
\end{equation}
where $\alpha$ denotes the weight and $f:X\mapsto\mathbb{R}^d$ is the function of extracting features from input, i.e., the encoder of detector. Then the gradients of incorporated features $f(\cdot):=f([X_{p_i},X_{a_i}])$ have:
\begin{equation}
    \begin{split}
        \frac{\partial \mathcal{L}}{\partial \theta_F} 
        &=\frac{\partial \mathcal{L}}{\partial f(\cdot)}\frac{\partial f(\cdot)}{\partial f(X_{p_i})}\frac{\partial f(X_{p_i})}{\partial \theta_F}+\frac{\partial \mathcal{L}}{\partial f(\cdot)}\frac{\partial f(\cdot)}{\partial f(X_{a_i})}\frac{\partial f(X_{a_i})}{\partial \theta_F} \\
        &=\alpha \frac{\partial \mathcal{L}}{\partial f(\cdot)}\frac{\partial f(X_{p_i})}{\partial \theta_F}+(1-\alpha)\frac{\partial \mathcal{L}}{\partial f(\cdot)}\frac{\partial f(X_{a_i})}{\partial \theta_F},
    \end{split}
\end{equation}
where $\frac{\partial f(X_{p_i})}{\partial \theta_F}$ and $\frac{\partial f(X_{a_i})}{\partial \theta_F}$ represent the gradients derived from the perturbed text $X_{p_i}$ and auxiliary information $X_{a_i}$, respectively. When the former introduced noises lead to the risk of trapping in local optima, the latter serves as momentum to alleviate this risk.

Next, we obtain auxiliary information and weights as follows.

(2) \textit{Implementation} The evasive tactics make the toxicity to be implicitly hinted so that the LLM needs to conduct in-depth analysis rather than just scratching the surface. Inspired by the Golden Circle principle~\cite{sinek2009start}, a structured thinking logic which understands things from phenomena to essence through multi-questioning, we introduce a \textit{How-Why-What} interrogation mechanism to require LLM to conduct considerable analysis as follows.

\textit{How Stage}: LLM explores \textit{how} to mine possible meaning and toxicity-related clues from perturbed text.

\textit{Why Stage}: According to the \textit{how} stage, LLM must give the reason \textit{why} it performs this procedure. 

\textit{What Stage}: LLM explores \textit{what} the possible meaning and toxicity-related clues is according to \textit{how} and \textit{why}.

Subsequently, we incorporate the auxiliary information into the perturbed text to enrich insights for detection. Since there are various differences (e.g., information gain) between the auxiliary information and the perturbed text in different samples, the information contributes differently to the toxicity judgment and we need to employ the auxiliary information in a dynamical way. Specifically, we introduce a difference-based dynamical cooperation, where we compute the point-wise difference between the features of both auxiliary information $\mathbf{x}_{a_{i},j}\in \mathbb{R}^d$ and perturbed text $\mathbf{x}_{p_i,j}\in \mathbb{R}^d$, and we then compute the dynamical weights as follows.
\begin{equation}
    \mathbf{w}_{p_i,j} = \sigma(\mathbf{W}_g^T(\text{CONV}((\mathbf{x}_{p_i,j}-\mathbf{x}_{a_{i},j})^2))+\mathbf{b}_g),
\end{equation}
where $\sigma$ represents the sigmoid function, and CONV denotes the convolutional operation to locally smooth the differences since the point-wise differences are sensitive due to outlier values introduced by perturbations. Then dynamical weights $\mathbf{w}_{p_i,j}$ are computed by global linear projection ($\mathbf{W}_g$ and $\mathbf{b}_g$), and we further incorporate $\mathbf{x}_{a_i,j}$ and $\mathbf{x}_{p_i,j}$ to get the cooperated features $\mathbf{f}_{p_i,j} \in \mathbb{R}^d$ as follows.
\begin{equation}
    \mathbf{f}_{p_i,j}=\frac{\mathbf{x}_{p_i,j}+\mathbf{w}_{p_i,j}\cdot \mathbf{x}_{a_i,j}}{(1+\mathbf{w}_{p_i,j})}
\end{equation}

\subsubsection{Discriminability Driven Feature Learning Strategy}

Noises in perturbations introduce trivial or irrelevant features, which renders the detector sensitive to various perturbed text. Hence, explicitly guiding the detector to learn critical features is essential for enhancing its robustness. Given that toxicity detection is inherently a classification task, the key to learning the critical feature lies in the core property of classification, i.e., discriminability. This is because the discriminability enables the detector to exhibit distinct probabilistic tendencies across classes (i.e., shape the robust inter-class classification boundary), making it the foundational support for breaking free from the interference of trivial features and enhancing the learning for critical ones. Accordingly, we propose a discriminability driven feature learning strategy to guide critical feature learning, which primarily comprises three stages:

(1) \textit{Feature Contribution Measuring}  To differentiate the discriminative power of features, we firstly quantify the probabilistic contribution caused by each feature on the class by attribution analysis based on integrated gradient as follows:
\begin{equation}
A_y(\mathbf{f}_{p_i,j})=(\mathbf{f}_{p_i,j}-\mathbf{f}^{\prime}) \int_0^1 \frac{\partial F_y(\mathbf{f}^{\prime}+\beta(\mathbf{f}_{p_i,j}-\mathbf{f}^{\prime}))}{\partial \mathbf{f}_{p_i,j}} \mathrm{d} \beta,
\end{equation}
where $\mathbf{f}^{\prime}\in \mathbb{R}^d$ denotes the reference features when there is no information input (set to zero vector by default). $F_y:\mathbb{R}^d\mapsto\mathbb{R}$ represents the projection from the features to the output probability of class $y$, and $\beta$ indicates the scaling coefficient. $A_y(\mathbf{f}_{p_i,j})\in \mathbb{R}^d$ denotes the quantified contribution of feature $\mathbf{f}_{p_i,j}$ for the output probability of class $y$. When $A_y^k(\mathbf{f}_{p_i,j})>0$ and $A^k_y(\mathbf{f}_{p_i,j})<0$, it indicates that the k-th element of feature $\mathbf{f}^k_{p_i,j}$ has a positive and negative impact, i.e., increasing and decreasing probabilistic tendencies on the label $y$, respectively. For example, increasing the feature $\mathbf{f}^k_{p_i,j}$ will make the detector output a larger probability on the class $y$ when $A^k_y(\mathbf{f}_{p_i,j})>0$ and vice versa.

(2) \textit{Discriminative Feature Differentiating} 
When one feature contributes positively or negatively to both the \textit{non-toxic} ($y=0$) and \textit{toxic} ($y=1$) classes, the feature is less discriminative since it cannot make the detector exhibit significant probabilistic differences between classes, contributing less in shaping the classification boundary. Hence, we differentiate the discriminability of one feature according to whether its attribution differs on classes as follows:
\begin{equation}
    \mathbf{m}^{less}_{\mathbf{f}_{p_i,j}}=\mathbf{I}\{A_{y=0}(\mathbf{f}_{p_i,j})A_{y=1}(\mathbf{f}_{p_i,j})>0\},
\end{equation}
where $\mathbf{I}$ is the indicator function. $\mathbf{m}^{less}_{\mathbf{f}_{p_i,j}}\in \mathbb{R}^d$ represents the mask for the less-discriminative elements within feature $\mathbf{f}_{p_i,j}$. The parts of $\mathbf{m}^{less,k}_{\mathbf{f}_{p_i,j}}=1$ contribute similarly on \textit{toxic} and \textit{non-toxic} classes. 

(3) \textit{Discriminability Driven Feature Learning} 
After differentiating discriminative power, we guide feature learning with discriminability as the core driver. First, we enhance the discriminative features to boost robustness. Specifically, when features rotate globally, the classification boundary should rotate accordingly, decoupling the discriminability from trivial attributes like absolute direction and position of the features and preserving it. Drawing on this insight, we rotate features and reclassify to strengthen discriminability. Given rotating high-dimensional features incurs high computational cost ($\mathcal{O}(d^2)$), we adopt a flipping (a special case with $\mathcal{O}(d)$ complexity) to rotate features, aiming to enhance the discriminability of features and improve robustness, shown as follows:
\begin{equation}
    \mathcal{L}_{more} = \text{CE}(\mathbf{W}_c^T(-\mathbf{f}_{p_i,j})+\mathbf{b}_c,y_{p_i,j}),
\end{equation}
where CE and $y_{p_i,j}$ are cross-entropy loss for classification and true label, respectively. $\mathbf{W}_c,\mathbf{b}_c$ denote the classifier head in the detector.

Furthermore, we unlearn less discriminative elements within features and relearn them to compose more robust features, shaping a reliable classification boundary. Considering neutral features should yield near-random predictions without any probabilistic tendency toward any class, we thus exploit a uniform distribution as supervision to unlearn the less discriminative elements, resetting them to be neutral. Subsequently, these elements are iteratively relearned multiple times to make them more robust.
\begin{equation}
    \mathcal{L}_{\text{less}} = \text{CE}(\mathbf{W}_c^T(\mathbf{f}_{p_i,j} \circ \mathbf{m}^{\text{less}}_{\mathbf{f}_{p_i,j}}) + \mathbf{b}_c, \mathbf{u}),
\end{equation}
where $\mathbf{u} \in \mathbb{R}^{|\mathcal{Y}|}$ is the uniform distribution over the label space.


\subsubsection{Historical Capability Replay Strategy}
To mitigate the forgetting of historical detection capability, we select top-k representative samples as memory samples to preserve the capability. Specifically, we mix memory samples into the training data, and we align the features of memory samples encoded by the old and current detectors since features encapsulate the detector’s capability as follows:
\begin{equation}
    \mathcal{L}_{align} =-\frac{1}{T}\sum_i^{T}(\frac{1}{N_{p_i}}\sum_j^{N_{p_i}}(\cos(\mathbf{f}_{p_i,j}^{old},\mathbf{f}_{p_i,j}^{cur})+Z_i)),
\end{equation}
where $Z_i=-\log\sum_k^{N_{p_i}}\exp(\cos(\mathbf{f}_{p_i,k}^{old},\mathbf{f}_{p_i,k}^{cur}))$ is the normalization factor. $\mathbf{f}_{p_i,k}^{old}$ and $\mathbf{f}_{p_i,k}^{cur}$ are the features of the j-th sample encoded by old and current detectors, respectively. $\cos$ is the cosine similarity. Finally, the total loss is:
\begin{equation}
    \mathcal{L}=\mathcal{L}_{cls}+\lambda(\mathcal{L}_{more}+\mathcal{L}_{less})+\gamma\mathcal{L}_{align},
\end{equation}
where $\mathcal{L}_{cls}=\text{CE}(\mathbf{W}_c^T\mathbf{f}_{p_i,j}+\mathbf{b}_c,y_{p_i,j})$ is the basic classification loss. $\lambda$ and $\gamma$ is the scale factors.





\section{Experiments}

\subsection{Experimental Setup}
\textbf{Dataset} Based on Jigsaw~\cite{jigsaw}, a widely used large-scale English dataset for toxicity detection, we construct a perturbed dataset by perturbing the Jigsaw with 9 types of perturbations (named DynEscape). In DynEscape, each type of perturbed text is separately type-wise managed, and every type owns 2584/1294/430 samples in train/test/valid splits. The perturbations include: \textit{\textbf{Insert}} (i\#dio!t), \textit{\textbf{Repeat}} (iiiddiott), \textit{\textbf{Swap}} (idito), \textit{\textbf{Remove}} (idot), \textit{\textbf{Homo}} (\textsf{id10t}), \textit{\textbf{Mask}} (id**t), \textit{\textbf{Abbr}} (abbreviation/slang replacement, bite me $\rightarrow$ BTM), \textit{\textbf{Auth}} (add authoritative text around the ordinary text to make it more convincing), and \textit{\textbf{Distract}} (add distracting words around the ordinary text to make it more confusing).
With DynEscape, we can evaluate the vulnerability of the existing detectors and static fine-tuning detection methods on evolving perturbed text, and we also can explore continual toxicity detection of ContiGuard against evolving perturbed text.
Furthermore, to examine the generalization and practical application of ContiGuard, we conduct a cross-dataset evaluation on NoisyHate~\cite{ye2025noisyhate} dataset, which is collected from mixed multi-types of human-written perturbed text in the wild.

\textbf{Evaluation Protocol} All our fine-tuned detectors are composed of an encoder of BERT-base and a classification head of a two-layer forward network. We conduct three types of experiments and use accuracy as the metric for classification. 




(1) We evaluate the exiting detectors on perturbed DynEscape to explore their vulnerability, including commercial APIs (PerspectiveAPI~\cite{lees2022new} and OpenAI Moderation~\cite{markov2023holistic}) and fine-tuned detectors (ToxicBERT/ToxicRoBERTa~\cite{hartvigsen2022toxigen}, Original/Multilingual/Unbiased variants of Detoxify~\cite{Detoxify}, Paradetox~\cite{logacheva2022paradetox}, and DeTox~\cite{dementieva2024overview}). 

(2) To analyze performance degradation of the statically fine-tuned detector, we fine-tune the detector with one specific perturbation type and test the detector on the remaining types to simulate the scenario where the detector continually encounters the types of perturbed text created over time. 

(3) We examine our continual learning (CL) strategies by comparing ContiGuard with adapted common CL methods on DynEscape, including LFL~\cite{jung2016less}, EWC~\cite{kirkpatrick2017overcoming}, LWF~\cite{li2017learning}, GEM~\cite{lopez2017gradient}, EPI~\cite{wang2023rehearsal}, CEAR~\cite{zhao2023improving}, DUCT~\cite{zhou2025dual} and SOYO~\cite{wang2025boosting}. \textit{Joint} trains the detector with all data, and \textit{Stream} sequentially optimizes the detector by pipeline data, and they act as the reference bound.

\begin{table*}
\centering
\caption{The results of ContiGuard and existing detectors on perturbed text. The subscripts indicate the relative performance degradation compared with ordinary text. The best scores are highlighted in \textbf{bold}.}
\label{tab:existing detection}
\begin{tabular}{lcccccccccc}
\toprule
Existing Detectors & Insert & Repeat & Swap & Remove & Homo & Mask & Abbr & Auth & Distract & Avg \\ \midrule
PespectiveAPI & 77.20 & 64.61 & 70.48 & 60.51 & 64.84 & 63.29 & 50.93 & 66.38 & 64.37 & $64.73_{25.9\%\downarrow}$ \\
OpenAI Moderation & 66.15 & 52.24 & 61.13 & 56.49 & 50.78 & 59.35 & 51.00 & 84.47 & 81.14 & $\text{62.53}_{25.2\%\downarrow}$ \\ \hline
ToxBERT & 60.90 & 60.97 & 58.81 & 58.34 & 53.55 & 55.87 & 58.35 & 73.26 & 75.12 & $\text{61.69}_{14.4\%\downarrow}$ \\
ToxRoBERTa & 50.93 & 50.39 & 51.24 & 52.32 & 50.15 & 50.31 & 51.39 & 55.26 & 79.83 & ${54.65}_{34.0\%\downarrow}$ \\
$\text{Detoxify}_{\text{Original}}$ & 61.59 & 63.76 & 61.90 & 57.73 & 56.96 & 59.27 & 56.03 & 71.87 & 75.66 & ${62.75}_{29.8\%\downarrow}$ \\
$\text{Detoxify}_{\text{Multilingual}}$ & 60.20 & 58.50 & 56.41 & 56.96 & 82.53 & 59.20 & 52.63 & 66.62 & 81.92 & ${63.88}_{26.6\%\downarrow}$ \\
$\text{Detoxify}_{\text{Unbiased}}$ & 51.31 & 51.00 & 53.32 & 53.40 & 50.15 & 51.93 & 51.55 & 78.59 & 80.14 & ${57.93}_{32.5\%\downarrow}$ \\
Paradetox & 51.31 & 51.47 & 51.85 & 51.62 & 49.92 & 52.63 & 51.16 & 69.55 & 72.57 & ${55.79}_{35.8\%\downarrow}$ \\
DeTox & 69.55 & 52.47 & 58.73 & 59.58 & 52.63 & 62.44 & 50.23 & 53.71 & 66.92 & ${58.48}_{28.3\%\downarrow}$ \\ \hline
ContiGuard & 85.57 & 87.23 & 85.18 & 83.97 & 90.40 & 82.94 & 80.58 & 92.19 & 90.15 & $\textbf{86.47}_{3.0\%\downarrow}$ \\
\bottomrule
\end{tabular}
\end{table*}

\begin{table*}
\centering
\caption{The results of ContiGuard and adapted CL methods. $\text{T}_i$ denote averaged performance of all orders at i-th moment. $\text{O}_{\text{T}_9}^j$ denote performance of j-th order at $\text{T}_9$ moment. * indicates significant difference at p<0.001.}
\label{tab: time-wise continue learning}
\begin{tabular}{lccccccccc|cccc}
\toprule
\begin{tabular}[c]{@{}l@{}}Continual\end{tabular}       & $\text{T}_1$ & $\text{T}_2$ & $\text{T}_3$ & $\text{T}_4$ & $\text{T}_5$ & $\text{T}_6$ & $\text{T}_7$ & $\text{T}_8$ & $\text{T}_9$ & $\text{O}_{\text{T}_9}^1$ & $\text{O}_{\text{T}_9}^2$ & $\text{O}_{\text{T}_9}^3$ & $\text{O}_{\text{T}_9}^4$ \\ \midrule
Joint      & 84.14  & 86.63  & 85.43  & 85.07  & 84.56  & 85.05  & 85.03  & 85.31   & 85.88  & 85.63 & 85.51 & 86.31 &  86.06      \\
Stream     & 84.23  & 84.64  & 76.52  & 77.36  & 78.17  & 79.55  & 79.73  & 74.28 & 78.26  &  79.49   &  74.08  & 78.95  &   80.49  \\ \hline
LFL        & 83.46        & 82.88        & 77.15        & 76.59        & 75.30        & 78.25        & 78.44        & 75.01        & 78.15        & 79.48  & 74.19   &  79.07   & 79.85       \\
EWC        & 84.66        & 84.88        & 76.41        & 78.48        & 77.89        & 80.89        & 78.64        & 76.01        & 79.32        &  79.81  & 75.43    & 80.25    & 81.80       \\
LWF        & 83.46        & 85.08        & 79.39        & 79.86        & 76.31        & 76.92        & 75.56        & 72.58        & 72.30        &  72.29    &  74.94   &  65.40    &  76.56    \\
EPI        & 84.12        & 81.46        & 81.60        & 79.13        & 79.90        & 78.23        & 76.98        & 78.26        & 79.20        &  79.03      &   79.50     &   79.19     &   79.07     \\
GEM        & 83.94        & 84.83        & 78.39        & 79.96        & 80.82        & 80.76        & 80.56        & 78.47        & 80.99        &  80.51      &   80.23     &   81.19     &   82.03     \\
CEAR       & 84.35        & 84.39        & 72.56        & 78.60        & 80.45        & 80.97        & 81.22        & 81.30        & 80.60        &  82.71      &   80.04     &   77.01     &   82.65     \\ 
DUCT        & 78.75 & 75.51 & 67.60 & 67.97 & 66.86 & 69.47 & 70.82 & 67.29 & 62.98  &  66.87      &   56.81     &   68.01     &   60.23     \\
SOYO       & 80.47 & 81.10 & 80.12 & 79.67 & 79.52 & 79.46 & 79.27 & 79.66 & 80.07 &   80.18      &   79.99     &   80.12     &   79.98     \\ \hline
ContiGuard & 87.17  & 89.27  & 86.20   & 86.42 & 85.41  & 86.29  & 85.41    & 84.92  & $\textbf{86.47}^*$   &  \textbf{85.67} & \textbf{86.28}  &  \textbf{86.34} &  \textbf{87.58}      \\ \bottomrule
\end{tabular}
\end{table*}

\subsection{Main Results}
Firstly, the results of existing detectors are shown in Tab.~\ref{tab:existing detection}. We find that the performance of existing detectors degrades significantly under all perturbations, with relative drops ranging from a maximum of 35.8\% to a minimum of 14.4\% compared to the ordinary text, denoting their susceptibility to perturbed text. This is because the perturbations obfuscate text to hide the toxicity, which makes the perturbed text significantly different from the ordinary ones, but the existing detectors are trained on ordinary text and ignore the adaptability to the perturbed text.

\begin{figure}
    \centering
    \includegraphics[width=0.95\linewidth]{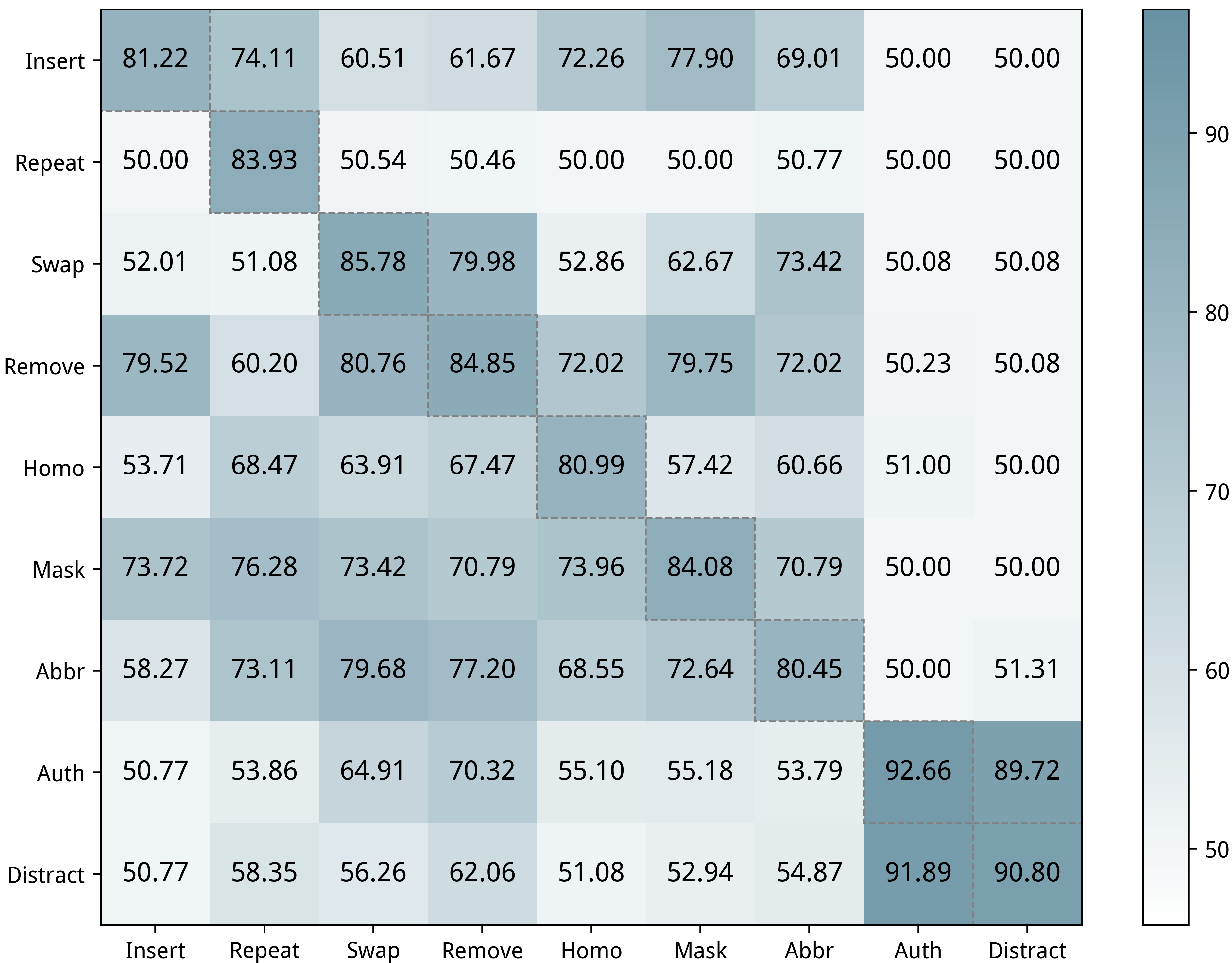}
    \caption{Results of statically fine-tuned detectors. Rows: types used for fine-tuning. Columns: types used for testing.}
    \label{fig:static finetuning}
\end{figure}

Secondly, the results of statically fine-tuned detectors are shown in Fig.~\ref{fig:static finetuning}. We find that the statically fine-tuned detectors perform well on the types of perturbed text used for fine-tuning (numbers on the diagonal), but perform poorly on other types of perturbed text (numbers outside the diagonal), where the performance drops in almost all types, and they even perform randomly on many types. We believe that the disparities among perturbations seriously impede the generalization across types so that the statically fine-tuned detectors on fixed specific types of perturbations are not suitable when the perturbations are continually created.

Finally, the continual detection results under different data orders are shown in Table~\ref{tab: time-wise continue learning}. All CL methods continually update the detection capability of the detector to capture continual adaptability to perturbations that occur over time. However, existing CL methods are limited because they never consider the unique properties of the continuous toxicity detection task and even perform worse than \textit{Stream}, while ContiGuard performs the best and outperforms \textit{Joint}, despite \textit{Joint} utilizing all data at every moment. We believe this is due to the hindrance of understanding the perturbed text due to semantic corruption and the difficulty in learning key features that are surrounded by the perturbation-introduced noisy features. Existing CL methods are never designed for these unique issues, resulting in limited performance. In contrast, ContiGuard enriches the perturbed text and performs robust feature learning to mitigate semantic corruption and noisy feature interference, achieving excellent and robust continuous detection.

\begin{figure}
    \centering
    \includegraphics[width=0.95\linewidth]{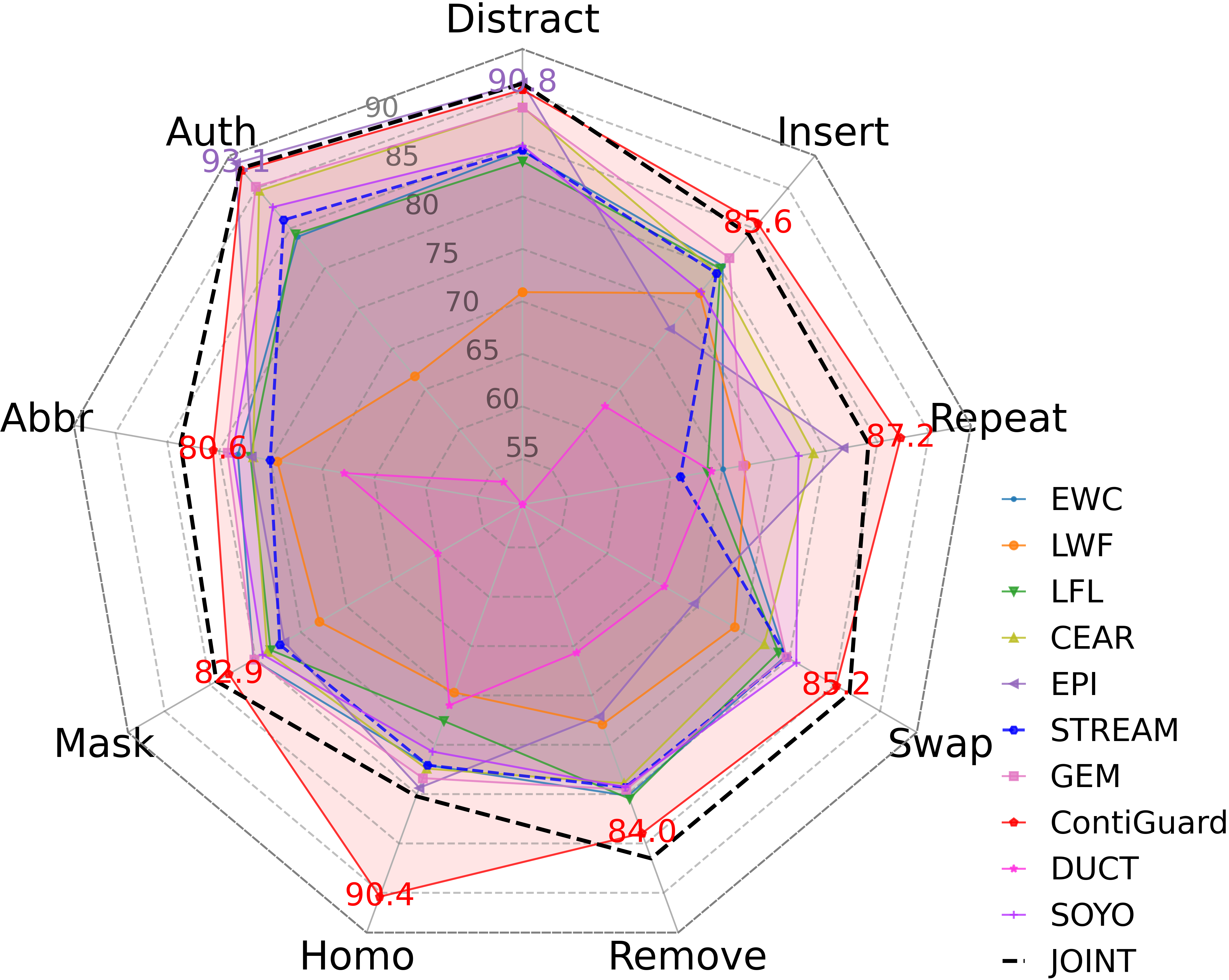}
    \caption{Results on different types of perturbed text at $T_9$. The numbers denote the best accuracy for each type.}
    \label{fig:rader}
\end{figure}

Furthermore, Fig.~\ref{fig:rader} shows the performances in every type of perturbed text at the $T_9$ moment. We find that ContiGuard performs the best in most perturbation types and is more balanced than existing CL methods across types. For example, despite EPI being close to ContiGuard in \textit{Auth} and \textit{Distract}, it works very poorly in \textit{Insert} and \textit{Swap}, indicating a serious imbalance. By contrast, ContiGuard is close to or exceeds \textit{Joint} on all perturbations.

\subsection{Analysis}

\textbf{Ablation Analysis}
To investigate the effectiveness of our proposed strategies, we conduct the ablated variants: (1) ablated LLM power: without LLM powered auxiliary information (w/o aux) and without dynamical cooperation (w/o coop); (2) ablated discriminability driven: without discriminability driven feature learning (w/o disc), without enhancement of more discriminative features (w/o more), and without suppression of less discriminative features (w/o less); (3) ablated historical replay: without memories (w/o mem) and without feature alignment (w/o align).

As Tab.~\ref{tab: time-wise ablation} shows, the performance of the variants drops to different degrees at most moments, and all drop significantly at the $T_9$ moment. These results prove that the proposed strategies are effective for continual toxicity detection against evolving perturbed text. Especially, the LLM powered enriching is helpful since the provided extra information enriches the insights of toxicity judgment.

\begin{table}[!htb]
\centering
\setlength{\tabcolsep}{1.3pt}
\caption{Ablation results at different moments.}
\label{tab: time-wise ablation}
\begin{tabular}{lccccccccc}
\toprule
\multicolumn{1}{c}{} & $\text{T}_1$ & $\text{T}_2$ & $\text{T}_3$ & $\text{T}_4$ & $\text{T}_5$ & $\text{T}_6$ & $\text{T}_7$ & $\text{T}_8$ & $\text{T}_9$ \\ \midrule
\multicolumn{1}{c}{ContiGuard} & 87.17 & 89.27 & 86.20 & 86.42 & 85.41 & 86.29 & 85.41 & 84.92 & \textbf{86.47} \\ \hline
 w/o aux & 83.11 & 84.63 & 80.52 & 81.03 & 80.03 & 81.85 & 82.12 & 80.06 & 81.97 \\
 w/o coop & 87.42 & 88.88 & 85.62 & 85.71 & 85.90 & 87.16 & 85.63 & 86.22 & 85.96 \\ 
\hline
 w/o disc & 86.20 & 87.71 & 84.08 & 85.63 & 86.18 & 85.30 & 84.49 & 84.74 & 85.71 \\
 w/o more & 86.36 & 88.84 & 85.61 & 86.07 & 86.22 & 85.78 & 84.75 & 84.59 & 85.89 \\
 w/o less & 86.78 & 88.86 & 86.49 & 86.11 & 86.17 & 85.59 & 84.44 & 84.06 & 86.25 \\ 
\hline
 w/o mem & 87.17 & 88.86 & 83.22 & 85.99 & 84.99 & 84.61 & 83.79 & 83.84 & 84.67 \\
 w/o align & 87.17 & 88.81 & 85.57 & 85.73 & 86.42 & 86.33 & 84.39 & 83.08 & 84.87 \\ 
\bottomrule
\end{tabular}
\end{table}

\begin{figure}
    \centering
    \includegraphics[width=0.9\linewidth]{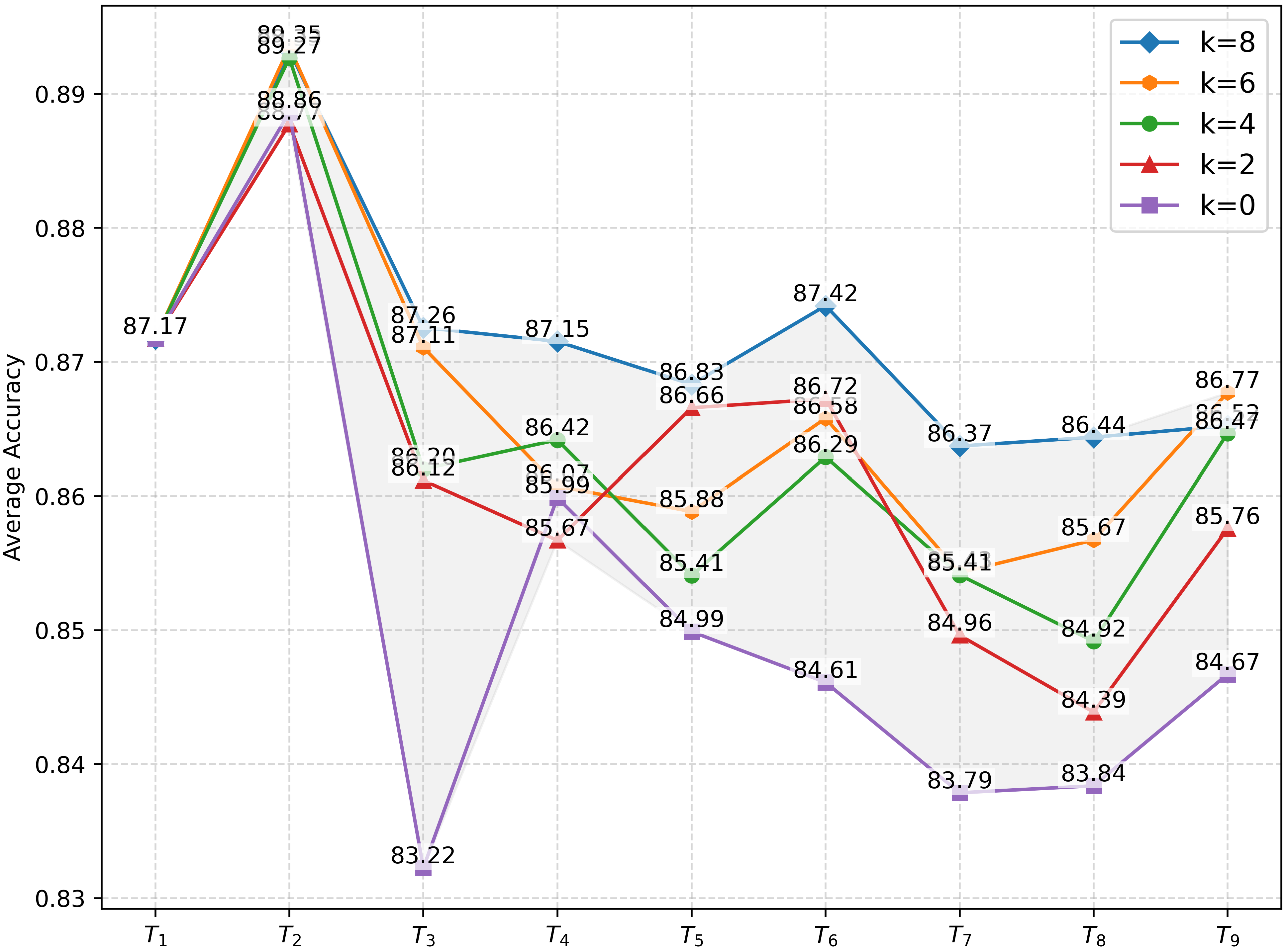}
    \caption{Results of different memory sample numbers.}
    \label{fig:memory size}
\end{figure}

\textbf{Analysis of the Memory Sample Number} To study the performance tendency under different numbers of memory samples, we experiment with varying numbers to observe the performance changes as Fig.~\ref{fig:memory size} shows. We find that under varying numbers, the performance change trend of toxicity detection is basically similar over time. In addition, as the number increases, the accuracies overall increase, and the performance degradation becomes relatively gentle. For example, when the number is 0, the performance decline trend is significant, while when it is 8, the decline trend becomes slower. We believe this is because memory samples can replay the detection capability to mitigate the forgetting.

\begin{figure}[!htb]
    \centering
    \includegraphics[width=\linewidth]{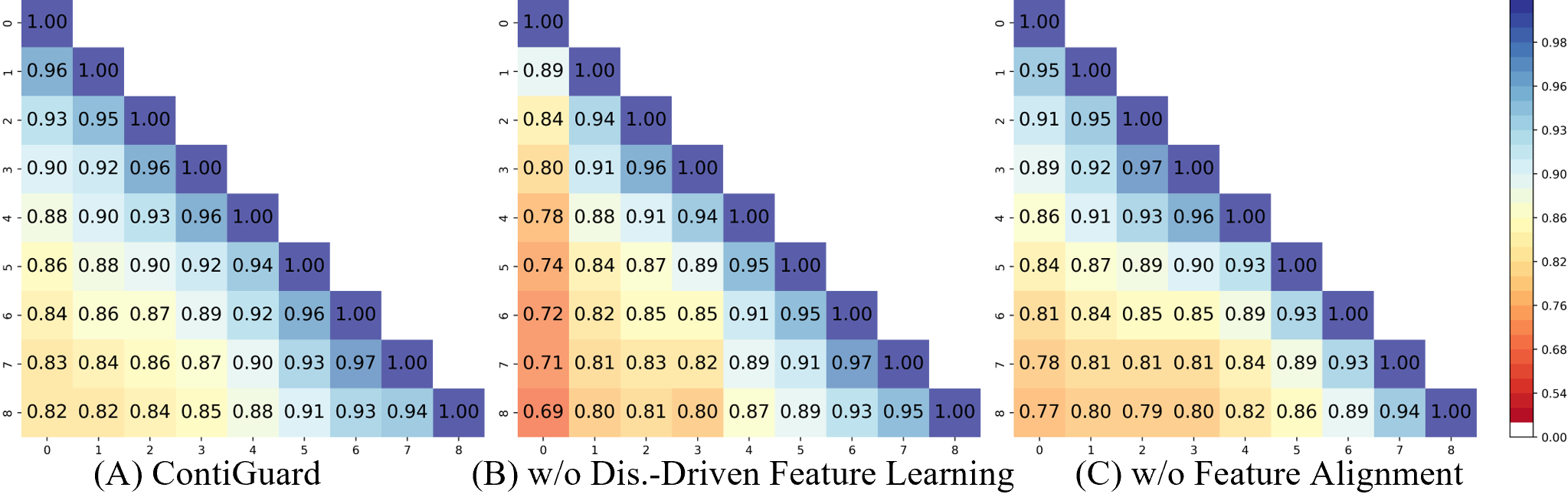}
    \caption{Retention rate of critical features, which denotes the proportion of historical critical features in current critical features. Each column shows the changes over time.}
    \label{fig: important features}
\end{figure}

\textbf{Retention Rate of Discriminative Features}
To examine the effectiveness of discriminability driven feature learning and historical capability replay, we analyze the retention rate changing of discriminative features after removing the two strategies. 
As Fig.~\ref{fig: important features} shows, the retention rates of discriminative features are all above 80\% over the moments in ContiGuard. However, when we ablate the two strategies, the retention rates decline to different degrees. Especially, the retention rate for perturbed text appearing at the $T_1$ moment significantly drops at the $T_9$ moment (see the first column), demonstrating that ContiGuard enables the detector to retain most of the learned discriminative features critical for toxicity detection, which mitigates the issue of capability forgetting.

\textbf{Analysis of Contribution Measurement} To explore different attribution methods, we compare integrated gradient (IG) and widely used SHAP~\cite{lundberg2017unified}, which resample output changes to estimate feature contribution. As Tab.~\ref{tab: contribution} shows, IG outperforms SHAP and we attribute it to the distinct responses of the two methods to subtle feature contribution changes induced by different perturbation manipulations: IG leverages gradients to capture such subtle changes sensitively, whereas SHAP resamples on the output for estimation, and this may attenuate the contribution of these subtle changes.

\begin{table}[!htb]
\centering
\setlength{\tabcolsep}{1.4pt}
\caption{Results of different contribution measuring methods.}
\label{tab: contribution}
\begin{tabular}{lccccccccc}
\toprule
\multicolumn{1}{c}{Methods} & $\text{T}_1$ & $\text{T}_2$ & $\text{T}_3$ & $\text{T}_4$ & $\text{T}_5$ & $\text{T}_6$ & $\text{T}_7$ & $\text{T}_8$ & $\text{T}_9$ \\ \midrule
IG & 87.17 & 89.27 & 86.20 & 86.42 & 85.41 & 86.29 & 85.41 & 84.92 & 86.47 \\
SHAP & 86.73 & 89.33 & 84.86 & 85.04 & 86.30 & 85.96 & 84.15 & 85.44 & 85.94 \\
\bottomrule
\end{tabular}
\end{table}

\begin{figure}[!htb]
    \centering
    \includegraphics[width=0.95\linewidth]{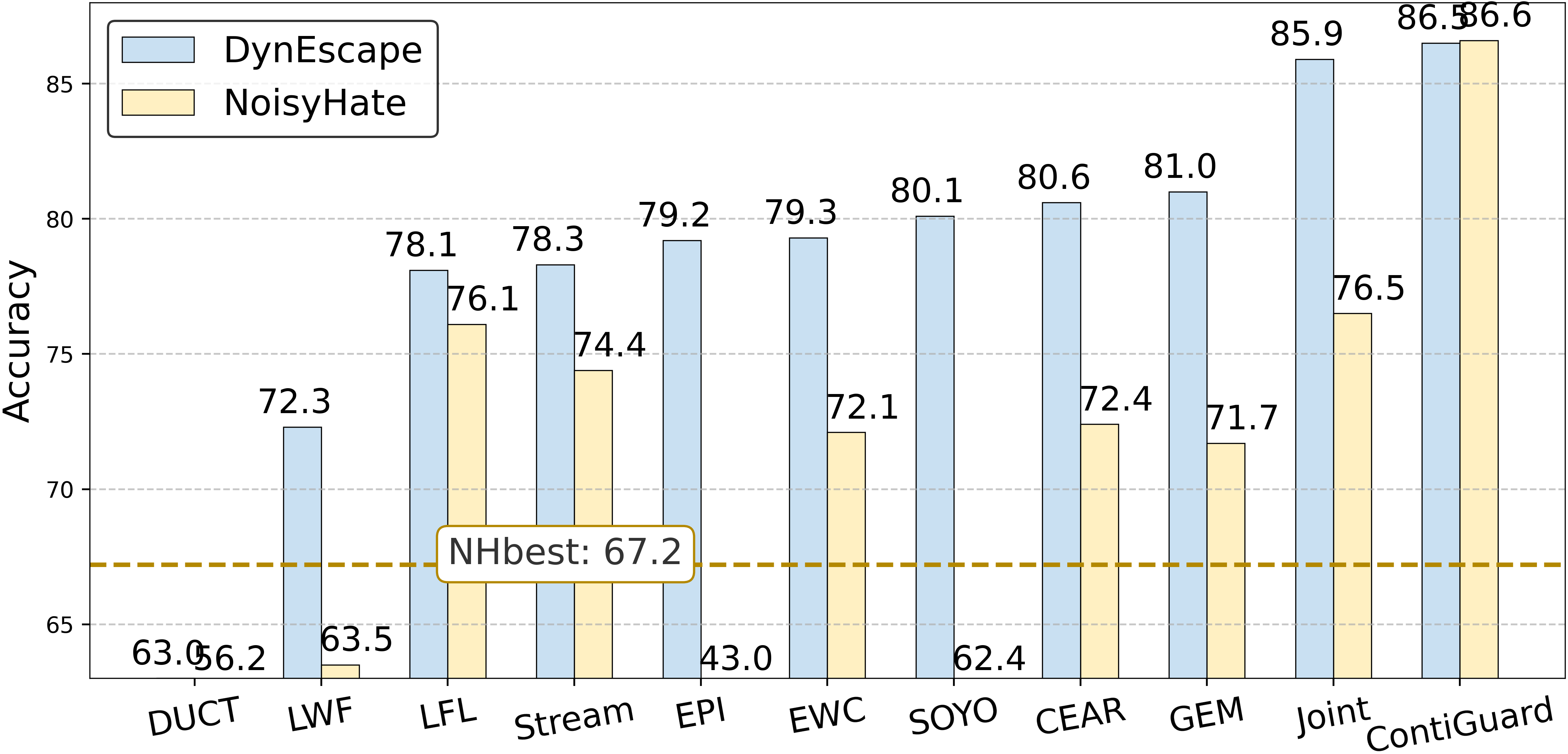}
    \caption{Results on DynEscape and NoisyHate.}
    \label{fig:cross test}
\end{figure}

\textbf{Application Evaluation}
To evaluate ContiGuard's practical application, we test CL methods on NoisyHate, as Fig.~\ref{fig:cross test} shows. NHbest is the original best detector (67.2)~\cite{ye2025noisyhate} on NoisyHate. We observe that CL methods mostly outperform NHbest, and ContiGuard exhibits the best performance and excellent cross-dataset adaptability. 
Furthermore, considering users' demand for costs, efficiency, and performance, we analyze from two perspectives: (1). Training with sampled datasets: we conduct training using sampled training data, with sampling proportions ranging from 20\% to 100\%. This analysis aims to quantify how the detector performance changes as training costs are lowered by reducing data usage. (2). Testing without LLM: we test the detector under a setting where LLM is excluded (referred to as $\text{ContiGuard}_{Eco}$). This setup is designed to explore how performance shifts when detection speed is enhanced by removing LLM during real application. 

\begin{figure}[!htb]
    \centering
    \includegraphics[width=0.95\linewidth]{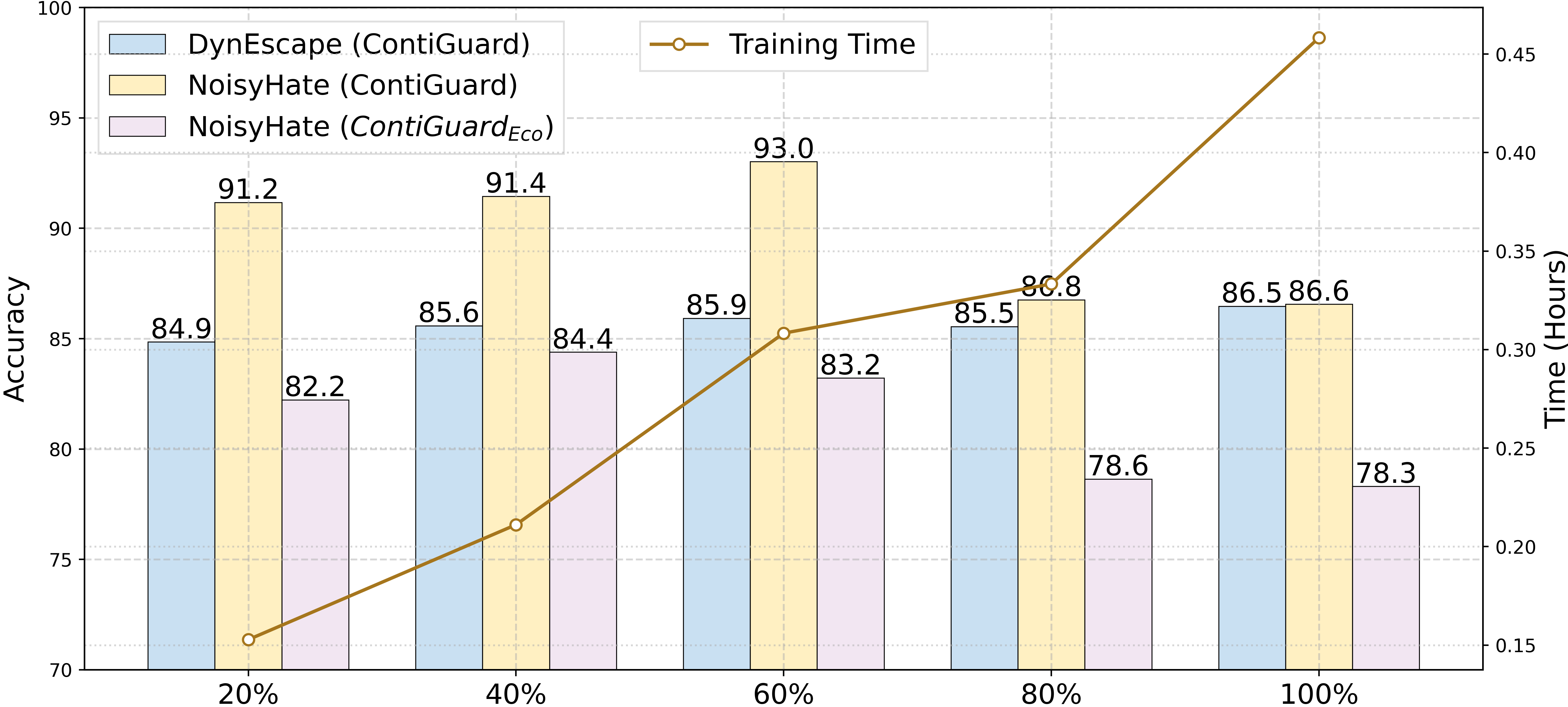}
    \caption{Results of different training data sampling rates.}
    \label{fig:sample rate}
\end{figure}

As Fig.~\ref{fig:sample rate} shows, a reduction in training data volume leads to a significant decrease in the training cost required per type of perturbation. For instance, when only using 20\% of training data, the training time per type of perturbation is reduced to approximately \textbf{\textit{0.15 hours}} (9 minutes), and ContiGuard still performs well on both the DynEscape and NoisyHate datasets. Furthermore, $\text{ContiGuard}_{Eco}$ delivers excellent performance on NoisyHate, with a detection speed about \textbf{\textit{220 items/s}}. This confirms $\text{ContiGuard}_{Eco}$ can balance detection performance and efficiency. Notably, we observe a trend: when training data is reduced to around 20\%–60\% of the original data volume, both ContiGuard and $\text{ContiGuard}_{Eco}$ show better performance on NoisyHate. We attribute this to differences in the perturbation characteristics of the datasets: DynEscape considers the situation of extreme perturbations, whereas NoisyHate is collected in the wild and owns milder perturbations, so less modeling on DynEscape makes it easier to generalize to NoisyHate. In addition, the consideration of extreme perturbations also enables ContiGuard trained on DynEscape to have a greater generalization margin when applied to real-world data like NoisyHate. For example, ContiGuard's performances on NoisyHate are all higher than those on DynEscape across proportions. In conclusion, users can tailor ContiGuard to their needs by adjusting two aspects: the volume of training data and whether to exploit LLM during testing, which allows a trade-off among cost, efficiency, and performance.

\begin{figure}[!htb]
    \centering
    \includegraphics[width=\linewidth]{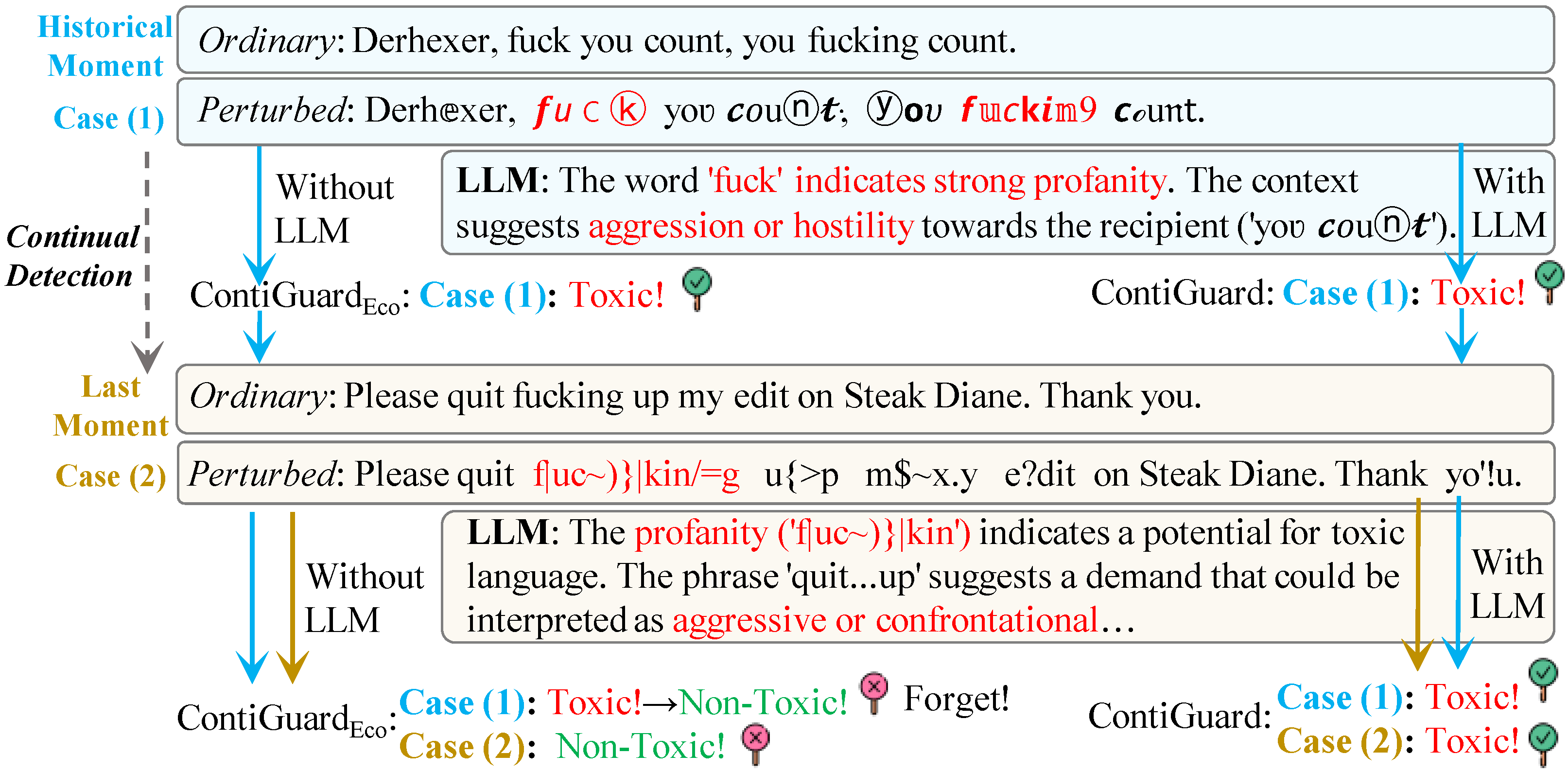}
    \caption{Case study for examining the impact of the LLM.}
    \label{fig:cases}
\end{figure}

\textbf{Case Study}
To examine the impact of LLM, we compare ContiGuard and $\text{ContiGuard}_{Eco}$ (excluding LLM) via cases in Fig.~\ref{fig:cases}.

Case (1) firstly appears at the historical moment and is obfuscated by homoglyph substitution. $\text{ContiGuard}_{Eco}$ and ContiGuard all identify the toxicity. However, when the last moment comes, after being trained on the text perturbed by character insertion, $\text{ContiGuard}_{Eco}$ fails to recognize the toxicity in case (1) since the disparity between perturbations breaks the historically learned detection ability. In contrast, ContiGuard enriches the perturbed text with the insights of aggressive attitude and the strong profanity `fuck' captured with LLM and learns from the insights to mitigate the ability breaking so that it still correctly detects the toxic case.

Case (2) firstly appears at the last moment and is perturbed by character insertion. $\text{ContiGuard}_{Eco}$ cannot associate and unify toxicity patterns in this case with the historically learned ones (e.g., perturbed `fuck') since the difference across perturbations obfuscates the detection. Hence, it fails to recognize the toxicity. ContiGuard can capture the insights of the aggressive attitude and the toxic term `fuck', which helps ContiGuard to perceive the similar toxic patterns from the insights in cases (1) and (2), enabling ContiGuard to successfully identify the toxicity in case (2).

Overall, ContiGuard uses LLM's analysis to enrich perturbed text to improve comprehension and to bridge different perturbations through shared insights derived from the analysis, which mitigates the forgetting of toxicity detection ability.

\section{Conclusion}
In this work, we first highlight the crucial yet overlooked challenge of continual toxicity detection against evolving evasive perturbed text drafted by malicious users, and we propose a novel ContiGuard framework exploiting strategies of LLM powered semantic enriching, discriminability driven feature learning, and historical capability replay to enable the detector to perform robust continual toxicity detection against evolving evasive perturbed text. Extensive experimental results show the superior performance of ContiGuard over the existing detectors and continual methods.

\begin{acks}
This work was supported by the grant from the National Natural Science Foundation of China (NSFC) project (Grant No. 62576256), the grant from Zhongguancun Academy (Grant No. 20240302), and the grant from the Key Laboratory of Computing Power Network and Information Security, Ministry of Education (Grant No. 2024ZD027).
\end{acks}

\bibliographystyle{ACM-Reference-Format}
\balance
\bibliography{sample-base}

\appendix

\section{Related Work}
\label{sec:related work}
\subsection{Toxicity Detection}
\label{sec:toxicity detection }
Various methods are proposed for toxicity detection, which can be primarily categorized into two types: ordinary text-focused and perturbations-augmented.

Ordinary text-focused methods are targeted at improving the performance on text without perturbations~\cite{bosco2023detecting,markov2023holistic,kebriaei2024persian,zhang2024efficient,pan2024umuteam,zhang2024don}. Many datasets across different languages have been constructed for this task~\cite{lashkarashvili2022toxicity,garg2023handling,delbari2024spanning}. Most of these datasets are curated from social platforms like X, Reddit, and Zhihu~\cite{sap2020social,haber2023improving,lu2023facilitating} with lots of manual efforts for annotation, and a few of them are generated by LLMs like GPT3~\cite{hartvigsen2022toxigen}.
The ordinary text-focused methods rely on lexical clues in normal text and are vulnerable to even simple attacks.

Perturbation-augmented methods enhance the detectors' robustness by incorporating perturbed text. They~\cite{le2022perturbations,bespalov2023towards,ye2025noisyhate} may manipulate characters, words, or sentences in the text, e.g., character swap and/or substitution, homoglyph/homophone substitution, decomposition, near-neighbor word replacement, and distract injection~\cite{kurita2019towards,cooper2023hiding,emmery2022cyberbullying,yu2024don}. The existing perturbation-augmented methods statically employ fixed specific types of perturbations to enhance the detectors' robustness, but they struggle to dynamically deal with increasing crafted perturbations over time.

The existing methods do not update the detector's capability so they cannot deal with the perturbed text that appears increasingly.
In contrast, we collect 9 types of perturbations and conduct continual learning for the detector when these perturbations emerge incrementally over time, enabling the detector to continually update its capability against increasing perturbations.

\subsection{Continual Learning}
\label{sec:continual learning}
Continual learning aims to continually learn from an incrementally appearing data stream~\cite{goodfellow2013empirical}. There are three incremental learning scenarios, i.e., domain-, class-, and task-incremental learning~\cite{van2022three}.

In domain-incremental learning, all tasks have the same label space but different data domains~\cite{michalski1986multi,tan2020incremental,garg2022multi,mirza2022efficient,wang2024multi}, e.g., \cite{zhou2025dual} proposes dual consolidation to create a representation space suitable for multiple domains incrementally and to merge the backbone of different stages, accommodating all seen domains throughout the learning. \cite{wang2025boosting} introduces a gaussian mixture compressor and domain feature re-sampler to store and balance prior domain data, and proposes a multi-level domain feature fusion network to extract domain feature, boosting the domain-incremental learning ability.

In task-incremental learning, the tasks are increasingly appearing and being solved~\cite{kanakis2020reparameterizing,douillard2020podnet,oren2021defense,gao2024enhancing,fu2024double}. Recently, \cite{lee2024incremental} introduces an adapter-based continual imitation learning framework to address the limitation of knowledge sharing by incrementally learning shareable skills from demonstrations, enabling sample-efficient task adaptation using the skills. \cite{yang2025eeg} employs wavelet packet transform to extract global-view spatial features via the regularized common spatial pattern and captures local-view spatial features via tangent space mapping from the Riemannian space to improve daily living of patients with strokes by task incremental learning.

In class-incremental learning, the number of categories is continually growing~\cite{wu2019large,zhang2020class,mittal2021essentials,zhu2021class,masana2022class,dong2022federated}. For instance, \cite{liu2024continual} employs wavelet transform to map the image into the frequency domain and balances the reusability and interference of output features based on the frequency domain similarity of the classes to mitigate the forgetting of previously acquired knowledge. \cite{he2024gradient} proposes to re-weight the gradients towards balanced optimization and unbiased classifier learning to address skewed gradient updates with biased weights. \cite{prompt25li} proposes to generalize conceptual knowledge learned from old classes to new classes by simulating human learning capability.

The continual toxicity detection problem falls under domain-incremental learning, where ordinary text is perturbed over time to be distributed in varying perturbation domains for evading detectors. To the best of our knowledge, there are no methods or datasets focusing on the continual toxicity detection problem, and our work is the first to explore the problem.

\section{Dataset Construction}
\label{sec:dataset}

Most existing toxicity datasets do not consider the perturbed text \cite{sap2020social,mathew2021hatexplain,hartvigsen2022toxigen,costa2024mutox}. Though a few datasets contain perturbed samples, there are only limited types of perturbed text produced by monotonous perturbation operations. Moreover, these datasets mix all types of perturbations together \cite{kirk2022hatemoji,ye2025noisyhate,cooper2023hiding},  by which we cannot examine the detectors' adaptability to changing perturbations. To address this issue, we construct \textit{DynEscape}, a perturbation-wise toxicity dataset covering 9 types of perturbations through three stages.

\textbf{Stage 1. \textit{Data Cleaning and Toxicity-relevant Word Selection}}. This stage aims to remove noises in raw data and find toxicity-relevant words so that we can perturb them to confuse detectors.

In this stage, we choose Jigsaw \cite{jigsaw} dataset as raw data since it contains lots of toxic text (223k). However, Jigsaw cannot be directly used for our purpose since it contains many noises. More importantly, perturbing toxicity-irrelevant words will have weak impact on the detector. Hence we perform data cleaning and toxicity-relevant word selection as the preprocessing.

\textbf{(1) Data Cleaning.} We clean noises in the Jigsaw, i.e., unknown words (identified by \textit{spellchecker} tool), private information (emails, user ids), and meaning-less text (repeated sentences less than 5 words). We also find that Jigsaw is seriously imbalanced between non-toxic and toxic samples, with a ratio approaching 10:1. Hence we re-sample the toxic and non-toxic samples in 1:1 ratio, and obtain 20k/20k toxic/non-toxic samples for perturbing.

\textbf{(2) Toxicity-relevant Word Selection.} We select words significantly influencing toxicity recognition. By perturbing these words, we wish to bypass detectors. Note that besides toxic words, many non-toxic words are also relevant to toxicity. Hence we propose four strategies to select toxicity-relevant word:

\textit{Online Toxic Words}. We collect online toxic words from Google, including bad words and the banned swear words (2.9K words).

\textit{Words Identified as Toxic}. We employ PerspectiveAPI \cite{lees2022new} to detect the toxicity of each word in the dataset. We collect the word identified as toxic (1.1K words).

\textit{Words Increasing Toxicity Scores}. We compute the expectation of decreased toxicity score (supplied by PerspectiveAPI) for each word to examine its contribution to the sample's toxicity. Specifically, given the word $w_i$ and the sentences $T_{w_i}$ it appears, we compute the expectation $E_{w_i}$ as:

\begin{equation}
    E_{w_i}=\frac{1}{|T_{w_i}|}\sum_{t_j\in T_{w_i}}[s (t_j)-s (t_j/w_i)],
\end{equation}
where $s (t_j)$ and $s (t_j/w_i)$ denote the toxicity score of the text $t_j\in T_{w_i}$ before and after removing $w_i$, and $s (t_j)-s (t_j/w_i)$ denotes the decreased toxicity score after removing $w_i$. The expectation $E_{w_i}$ represents the contribution of the word $w_i$ to the toxicity of text. We select the words whose expectations are greater than 0 (1.5K words).

\textit{Spuriously Correlated Words}. Some words often appear with toxic label so that detectors tend to identify the text including these words as toxic, i.e., the words are spuriously correlated to labels~\cite{garg2023handling}. We compute mutation information to get such words \cite{zhang2023mitigating}.
\begin{equation}
    MI = \frac{p (w_i,c)}{p (w_i,\cdot)p (\cdot,c)},
\end{equation}
where $p (w_i,\cdot)$ and $p (\cdot,c)$ are the marginal distribution of the word $w_i$ and the label $c$, and $p (w_i,c)$ is the joint distribution of $w_i$ and $c$. The larger $MI$, the stronger the spurious correlation between $w_i$ and $c$. Based on multiple manual checks and previous empirical setting~\cite{zhang2023mitigating}, we set the spurious correlation threshold as the sum of the mean and standard deviation of $MI$ (1.1K words).

With these strategies, we totally collect 5.6K toxicity-relevant words after de-duplicating.

\textbf{Stage 2}. \textit{Perturbation Tool Creation}. This stage aims to provide the definitions of perturbations and then create a perturbation tool, which will be used as the attacker to perturb samples.

In this stage, we create a perturbation tool that could perform 9 types of perturbations. 

\textbf{(1) Perturbation Preparation} We prepare necessary material for perturbation, including 2.7K homoglyphs, 80.6K abbrs/slangs, and 3.3K roles.

\textbf{(2) Perturbation Patterns} We then define perturbation patterns as follows (Tab.~\ref{tab:appendix pert ways} shows the examples).

\textit{Character-level Perturbation}. \textbf{Insert} randomly adds special characters (from Python \emph{string} package) into the text. \textbf{Remove}/\textbf{Repeat} randomly deletes/duplicates characters. \textbf{Swap} randomly exchanges the order of characters. \textbf{Homoglyph} randomly replaces characters with visually similar characters (using prepared 2.7K homoglyphs).

\textit{Word/Phrase-level Perturbation}. \textbf{Maskword} randomly obscures characters in the selected words with special characters. \textbf{Abbreviation} employs abbreviations/slangs to replace the associated words or phrases (using prepared 80.6K abbrs/slangs).

\textit{Sentence-level Perturbation}. \textbf{Distract} adds the prefix consisting of extra random non-toxic words before the text. \textbf{Authorization} firstly instructs the LLMs to generate a self-introduction by acting as authoritative experts (using prepared 3.3K roles). Then, it adds the introduction before text to confuse detectors. This is an exploratory perturbation for evading detectors since the emergence of LLMs.

Based on the definitions, we implement 9 perturbations to create a perturbation tool that provides automated perturbation patterns. 

We finally conduct a human verification by manually checking 50 perturbed samples for each perturbation type (450 in total). Results show that all  perturbed samples conform to the definitions, which proves  our tool's reliability.

\begin{table}[!htb]
\centering
\caption{Examples of different perturbations.}
\label{tab:appendix pert ways}
\begin{tabular}{ll}
\toprule
\textbf{Perturbations} & \textbf{Examples} \\
 \midrule
Insert & moron$\rightarrow$mo*ro\#n; idiot$\rightarrow$idi+ot \\
Remove (Mv) &  moron$\rightarrow$moro; idiot$\rightarrow$idot\\
Repeat &  moron$\rightarrow$ mooron; idiot$\rightarrow$ iddiot\\
Swap & fool$\rightarrow$ folo; idiot$\rightarrow$ idoit\\
Homoglyph (Homo) &  idiot$\rightarrow$\textsf{id10t}; fool$\rightarrow$ f001
\\
Maskword (Mask) & fool$\rightarrow$f**l; idiot$\rightarrow$i*iot \\
Abbreviation (Abbr) & fuck$\rightarrow$ 4Q; bite me$\rightarrow$ BTM\\ 
Distract (Dis) & [text]$\rightarrow$[apple earth...];[text] \\
Authorization (Auth) & [text]$\rightarrow$[I am a scientist...];[text] \\
 \bottomrule
\end{tabular}
\end{table}

\textbf{Stage 3}. \textit{Adversarial Attack and Quality Assessment}. This stage aims to attack detectors using the samples perturbed by our tool and to select the challenging perturbed samples to compose \textit{DynEscape}.

In this stage, we utilize the perturbation tool as an attacker to disturb clean samples for evading detectors (e.g., safety-aligned LLama3). The perturbed samples bypassing detectors are selected to compose our \textit{DynEscape}. Note that to prevent detectors from taking shortcuts by judging whether the text has been disturbed, non-toxic samples are also perturbed in the same way as toxic ones.

We collect all successful samples to evade detectors, and they are reallocated with 4K non-overlapped samples per perturbation. We then split them into train/valid/test subsets with a ratio of 6:1:3 (2584/430/1294). Finally, our perturbation-wise dataset \emph{DynEscape} consists of 38K ((2584+430+1294)*9) challenging perturbed samples.

To ensure the quality of our dataset, we randomly select 50 samples per perturbation and employ three master students to evaluate the semantic consistency before/after perturbations on a scale of 0 to 1. We get a score of $\text{0.88}_{\pm \text{0.07}}$ averaged over all samples and more than two masters assign a score exceeding 0.5 in 99\% of the samples, indicating a high consistency of our dataset.

During data collection, we employ online toxic words collected by Google, which are available by `https://github.com/coffee-and-fun/google-profanity-words/tree/main', and we set perturbation rate to 20\%, i.e., 20\% of the words in the original text are perturbed. We utilize punctuation as special characters from the Python package `\emph{string}', and we collect homoglyphs from the `\emph{homoglyphs}' python tool to replace the characters. In addition, we collect the abbreviations/slangs from the online resources, including `onlineslangdictionary.com',`slang.net', `www.acronymfinder.com' and `acronymsandslang.com'.

\section{Experimental}
\label{sec:experimental}
\subsection{Details of Experimental Implementation}
In all experiments, we use the bert-base-uncased model (110M) as the encoder of the detector to encode the sentences, where the maximum sequence length is set to 360. The parameters of the models are updated using the AdamW optimizer with a learning rate of 2e-5. The random seed for all experiments is set to 0, and training is conducted with an early stopping strategy. All experiments are conducted on the A800 GPU. Considering the strong reasoning ability, we employ gpt-4o-mini to capture the auxiliary information.

\subsection{Introduction of Existing Methods}
LFL \cite{jung2016less} reduces knowledge forgetting by regularizing the parameters between old and new classifiers. 

EWC \cite{kirkpatrick2017overcoming} uses the Fisher matrix to assess parameter importance and applies weighted regularization based on that importance. 

LWF \cite{li2017learning} transfers knowledge by having the old model generate pseudo-labels for the new model. 

GEM \cite{lopez2017gradient} stores gradient information from past tasks and constrains the gradient direction of the current task to prevent it from conflicting with the gradient directions of old tasks.

EPI \cite{wang2023rehearsal} trains task-specific prefix parameters for each task and identifies the task IDs of test samples to select the appropriate prefix parameters for classification. 

CEAR \cite{zhao2023improving} learns memory-insensitive prototypes and uses memory augmentation to reduce overfitting and enhance performance.

DUCT~\cite{zhou2025dual} proposes dual consolidation to construct a unified representation space applicable to multiple domains and integrate backbones from different moments to mitigate the forgetting.

SOYO~\cite{wang2025boosting} introduces a gaussian mixture compressor and feature re-sampler to store balance prior domain data, and proposes a multi-level feature fusion network to enhance domain feature extraction, boosting the domain-incremental learning ability.

\subsection{Supplementary Experiments}



\textbf{The performance of directly using gpt-4o-mini} We also explore how 4o-mini performs on perturbed text, as Tab.~\ref{tab:4o-mini} shows. We find that 4o-mini performs well on ordinary text (DynEscape (ord.)), but its performance still degrades to 69.50 with a drop of $19\%$ when dealing with perturbed text (DynEscape). In addition, compared to DynEscape, 4o-mini performs better on NoisyHate due to minor perturbations within NoisyHate, but it was still worse than ContiGuard and even $\text{ContiGuard}_{Eco}$. We believe that using 4o-mini's reasoning capability to extract auxiliary information to enhance the detector is a good alternative approach compared to directly using it for toxicity detection against perturbed text.

\begin{table}[!htb]
\centering
\caption{Result of 4o-mini. DynEscape (ord.) represents the original ordinary sample corresponding to DynEscape.}
\label{tab:4o-mini}
\begin{tabular}{lccc}
\toprule
        & DynEscape (ord.) &   DynEscape       & NoisyHate \\ \midrule
4o-mini &      85.81       &  $69.50_{19\%\downarrow}$  &   80.06       \\ \bottomrule
\end{tabular}
\end{table}

\end{document}